%% file: main.tex
\documentclass[10pt,twocolumn,letterpaper]{article}

\usepackage[pagenumbers]{main}

\usepackage[table]{xcolor}
\usepackage{microtype}
\usepackage{graphicx}
\usepackage{booktabs}
\usepackage{amsmath}
\usepackage{amssymb}
\usepackage{mathtools}
\usepackage{amsthm}

\usepackage{color}
\usepackage{enumitem}
\usepackage{fontawesome5}
\usepackage{multirow}
\usepackage{makecell}

\usepackage{algorithmic}
\usepackage{algorithm}
\usepackage{etoolbox,siunitx}

\usepackage{colortbl}

\definecolor{lblue}{RGB}{231, 66, 52}

\usepackage[pagebackref=false,breaklinks,colorlinks,citecolor=lblue]{hyperref}

\def\paperID{1}
\def\confName{RoboSense}
\def\confYear{IROS 2025}

\title{Learning to Navigate Socially Through Proactive Risk Perception}

\author{Erjia Xiao\thanks{Equal contribution.}\\
HKUST (GZ)\\
{\tt\footnotesize exiao469@connect.hkust-gz.edu.cn}
\and
Lingfeng Zhang\footnotemark[1]\\
Tsinghua University\\
Xiaomi EV \\
{\tt\footnotesize zlf25@mails.tsinghua.edu.cn}
\and
Yingbo Tang\\
Institute of Automation, CAS\\
{\tt\footnotesize tangyingbo2020@ia.ac.cn}
\and
Hao Cheng\\
HKUST (GZ)\\
{\tt\footnotesize hcheng046@connect.hkust-gz.edu.cn}\\
\and
Renjing Xu\\
HKUST (GZ)\\
{\tt\footnotesize renjingxu@hkust-gz.edu.cn}\\
\and
Wenbo Ding\\
Tsinghua University\\
{\tt\footnotesize ding.wenbo@sz.tsinghua.edu.cn}\\
\and
Lei Zhou, Long Chen, Hangjun Ye, Xiaoshuai Hao\thanks{Project leader.}
\\
Xiaomi EV\\
{\tt\footnotesize  \{zhoulei21, chenlong37, yehangjun, haoxiaoshuai\}@xiaomi.com}\\
}

\begin{document}

\maketitle

\input{sections/0_abstract}

\input{sections/1_intro}

\input{sections/2_related_work}
\input{sections/3_method}

\input{sections/4_experiments}
\input{sections/5_conclusion}

{
\small
\bibliographystyle{ieeenat_fullname}
\bibliography{main}
}

\end{document}

%% file: sections/0_abstract.tex
\begin{abstract}
In this report, we describe the technical details of our submission to the IROS 2025 RoboSense Challenge Social Navigation Track. This track focuses on developing RGBD-based perception and navigation systems that enable autonomous agents to navigate safely, efficiently, and socially compliantly in dynamic human-populated indoor environments. The challenge requires agents to operate from an egocentric perspective using only onboard sensors including RGB-D observations and odometry, without access to global maps or privileged information, while maintaining social norm compliance such as safe distances and collision avoidance. Building upon the Falcon model, we introduce a \textbf{Proactive Risk Perception Module} to enhance social navigation performance. Our approach augments Falcon with collision risk understanding that learns to predict distance-based collision risk scores for surrounding humans, which enables the agent to develop more robust spatial awareness and proactive collision avoidance behaviors. The evaluation on the Social-HM3D benchmark demonstrates that our method improves the agent's ability to maintain personal space compliance while navigating toward goals in crowded indoor scenes with dynamic human agents, achieving \textbf{2nd place} among 16 participating teams in the challenge.
\end{abstract}

%% file: sections/1_intro.tex
\section{Introduction}
\label{sec:intro}

\begin{figure}[ht!]
    \centering
    \includegraphics[width=1.0\linewidth]{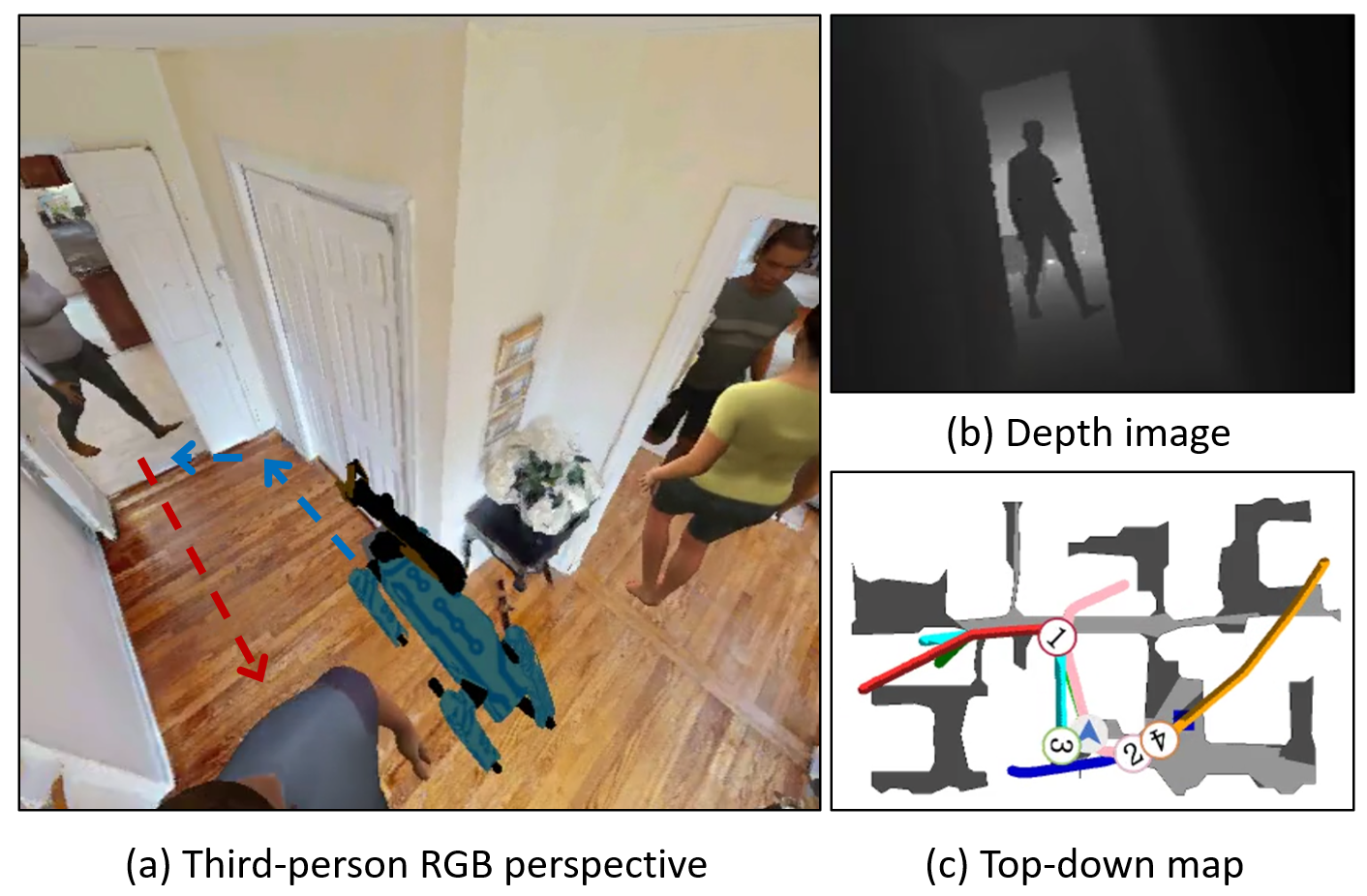}
    \caption{\textbf{SocialNav task illustration, adapted from~\cite{gong2025cognition}.} In (a), the robot navigates toward a goal (blue dashed lines) while predicting human trajectories (red dashed line) and avoiding them. The robot uses depth input as shown in (b). (c) offers a top-down map for reference, which is not used by the robot.}
    \label{fig:task_illustration}
\end{figure}

\noindent With the advancement of embodied intelligence~\cite{zhang2025vtla, tang2025affordgrasp}, Social Navigation (SocialNav) denotes the capability of autonomous robots to navigate in human-shared environments while adhering to social conventions and maintaining culturally appropriate interpersonal distances~\cite{mavrogiannis2023core}.
Unlike conventional navigation paradigms that depend on pre-constructed static maps, SocialNav necessitates continuous adaptation to dynamic human behaviors while ensuring both collision-free trajectories and socially acceptable conduct. SocialNav Task is illustrated in Figure~\ref{fig:task_illustration}.

Contemporary approaches to this challenge predominantly employ reinforcement learning (RL) frameworks~\cite{kapoor2023socnavgym,wang2024multi,hirose2024selfi}.
However, conventional RL-based methods frequently exhibit myopic behavior, focusing on immediate obstacle avoidance rather than anticipating future scenarios~\cite{chen2019crowd,li2019sarl}.
While hierarchical architectures combining global path planners with local RL policies have shown promise~\cite{perez2021robot}, they inherently depend on comprehensive environmental knowledge~\cite{cancelli2023exploiting}, rendering them impractical for real-world deployment where such information is unavailable.
The integration of human trajectory forecasting has demonstrated substantial benefits for dynamic collision avoidance~\cite{liang2019peeking,nishimura2020risk}, though primarily in outdoor autonomous driving contexts~\cite{liang2020garden,liang2020simaug}.
Indoor environments present distinct challenges, including constrained maneuvering space and elevated collision probabilities~\cite{perez2021robot}.
The recently proposed Falcon framework addresses these limitations through future-aware mechanisms, incorporating a Social Cognition Penalty and Spatial-Temporal Precognition Module to enhance trajectory prediction capabilities.

In the IROS 2025 RoboSense Challenge Social Navigation Track, participants are tasked with developing navigation policies based exclusively on egocentric RGBD observations and odometry data.
The benchmark dataset, Social-HM3D~\cite{gong2025cognition}, provides photo-realistic large-scale indoor scenes populated with collision-aware human agents, offering a rigorous testbed for evaluating social navigation algorithms. Building upon the Falcon~\cite{gong2025cognition}, we introduce a \textbf{Proactive Risk Perception Module} that enhances the agent's capability to assess and respond to collision risks in human-populated environments, as illustrated in Figure~\ref{fig:meth_overview}.
Our module employs a neural network to predict continuous risk scores for each nearby human based on their relative distances, with graduated risk levels spanning from safe zones to immediate danger regions.
By incorporating distance-weighted loss functions that emphasize high-risk scenarios, our approach enables more nuanced spatial reasoning and anticipatory collision avoidance strategies.

%% file: sections/2_related_work.tex
\section{Related Work}
\label{sec:related_work}

\textbf{Social Navigation.} The SocialNav paradigm~\cite{perez2021robot,francis2023principles} extends traditional point-goal navigation by incorporating dynamic human agents into navigation scenarios, first formalized in the iGibson SocialNav Challenge~\cite{xia2020interactive}.
Early implementations featured simplified human representations with constrained motion patterns.
Contemporary research benefits from advanced simulation platforms such as Habitat 3.0~\cite{puig2023habitat}, which provides photorealistic rendering and naturalistic human animations.
The SocialNav domain has attracted extensive research attention across robotics, computer vision, and behavioral studies~\cite{singamaneni2024survey, moller2021survey, zhang2025mapnav, zhang2025multi, wu2025evaluating, zhang2024trihelper, wu2025evaluating, zhang2025nava}.
Foundational work in multi-agent collision avoidance~\cite{van2011reciprocal, rvo, learned} and dynamic obstacle navigation~\cite{mpdyn} has evolved to specifically address human-robot interaction challenges~\cite{guzzi2013human, socialforces, sa-cadrl, li2025seeground, li2025_3eed, xu2025lima}.
Several approaches model interpersonal dynamics through spatio-temporal graph representations~\cite{social-graph, socialattention} to capture evolving agent relationships.
Recent investigations emphasize egocentric navigation in realistic settings~\cite{rudenko2020thor, martin2021jrdb, vuong2023habicrowd, chen2023clip2Scene, chen2023towards}.
Our work distinguishes itself by incorporating \textbf{Proactive Risk Perception} that complements trajectory prediction capabilities for enhanced collision avoidance.

\textbf{Auxiliary Learning in Navigation.} Auxiliary task learning has proven effective for improving task performance in navigation~\cite{mirowski2016learning}.
Diverse auxiliary objectives have been explored~\cite{jaderberg2016reinforcement,lin2019adaptive}, including environmental property prediction and agent state forecasting~\cite{agrawal2015learning, jaderberg2016reinforcement}.
Recent work has leveraged privileged information such as human-agent spatial relationships as auxiliary signals~\cite{cancelli2023exploiting}.

\textbf{Trajectory Prediction.}
The forecasting of trajectories plays a fundamental role in autonomous system safety~\cite{rudenko2020human,huang2023multimodal,hao2021matters,hao2022listen,hao2021matters,hao2023dual,hao2023uncertainty,hao2024mbfusion,hao2025mapfusion,hao2024your,hao2025safemap,kong2025largead,kong2023robo3d,xie2023robobev, survey_3d_4d_world_models}.
Classical methodologies include physics-based models such as Social Force~\cite{socialforces}, which simulate social dynamics through force interactions.
Contemporary approaches span three paradigms: physics-based methods deriving from Newtonian mechanics~\cite{elnagar2001prediction, zernetsch2016trajectory}, data-driven techniques learning from historical patterns~\cite{alahi2016social}, and intent-reasoning methods that model goal-directed behavior~\cite{vasquez2016novel}.

%% file: sections/3_method.tex
\section{Methodology}
\label{sec:method}

\begin{figure*}[t]
    \centering
    \includegraphics[width=1.0\linewidth]{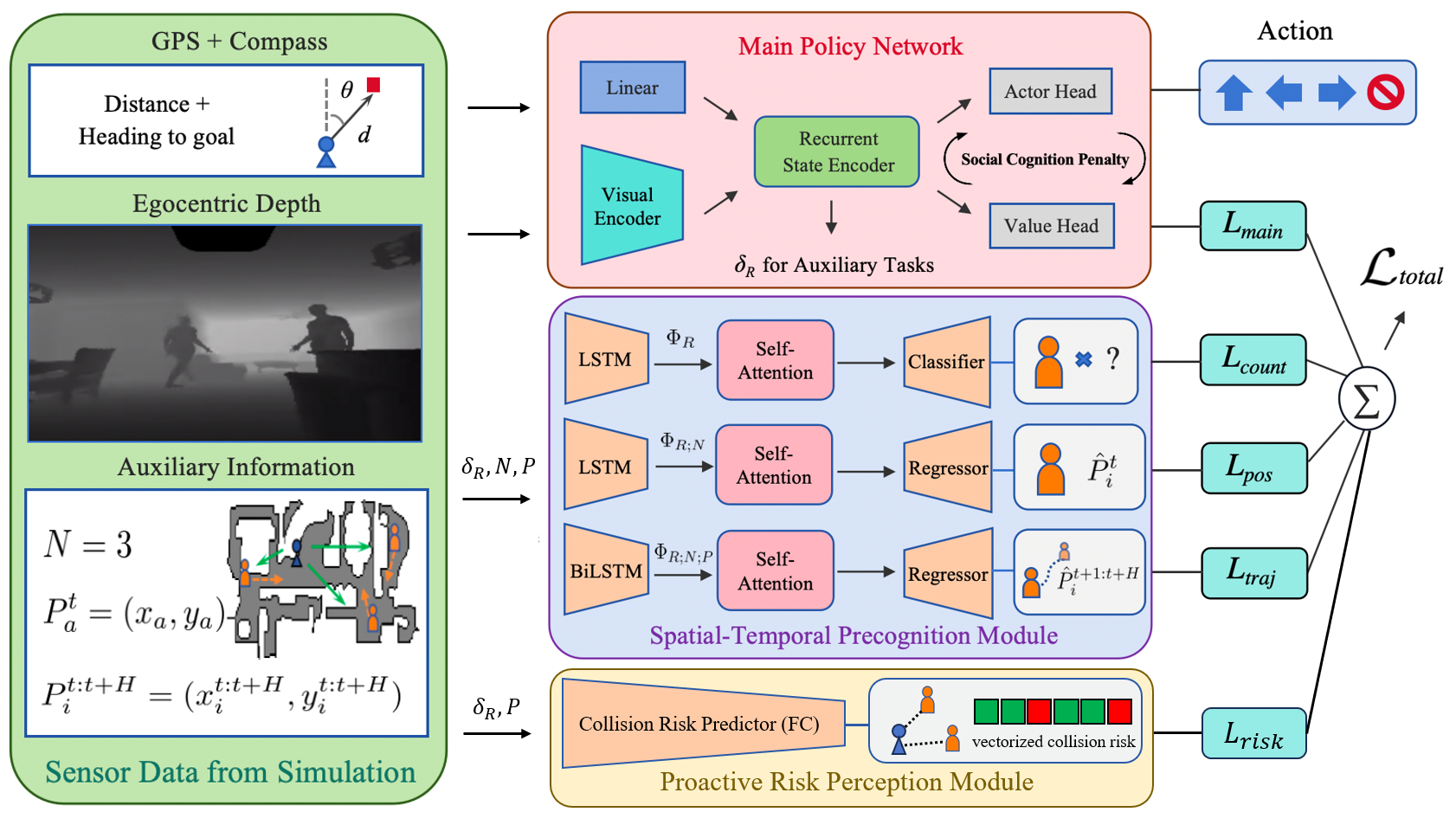}
    \caption{\textbf{Method overview, adapted from~\cite{gong2025cognition}.} Building upon the Falcon framework, our approach integrates a \textbf{Proactive Risk Perception Module} that operates alongside the main policy network. The policy network processes depth and GPS+Compass observations, guided by social cognition penalties. During training, the state encoder outputs are fed to both Falcon's original Spatial-Temporal Precognition Module and our proactive risk perception module. The risk module predicts distance-based collision risks for each nearby human, generating an auxiliary loss that enhances the agent's spatial awareness and collision avoidance capabilities.}
    \label{fig:meth_overview}
\end{figure*}

\subsection{Task Definition}
We formulate the social navigation challenge as follows: an autonomous agent $a$ operates within an environment containing $N$ dynamic human entities, indexed as $i \in \{1, \dots, N\}$. Beginning from the initial state $q_a \in Q$, the agent must generate a trajectory $\tau_a$ toward the target state $g_a \in Q$ while maintaining collision-free paths with both static structures and moving humans. This objective can be expressed as:

\begin{equation}
\begin{aligned}
\tau_a &= \arg \min_{\tau \in \mathcal{T}} \left( c_a(\tau) + \lambda_a c_a^s(\tau, {\tau}_{1:N}) \right) \\
\text{s.t.} \quad & A_a(\tau_a) \notin C_{\text{obs}},\quad A_a(\tau_a) \cap A_i(\tau_i) = \emptyset, \\
& \tau_a(0) = q_a, \quad \tau_a(T) = g_a.
\end{aligned}
\end{equation}

Here, $c_a$ denotes the trajectory cost toward goal completion; $c_a^s$ represents social compliance costs; $A(\tau)$ indicates spatial occupancy along trajectory $\tau$; $C_{\text{obs}}$ designates static environmental obstacles; $T$ marks episode termination; and $\lambda_a$ serves as a balancing coefficient. The constraints ensure collision-free navigation until goal achievement.

\subsection{Falcon Framework Overview}

Our method builds upon the Falcon architecture~\cite{gong2025cognition}, which comprises a main policy network trained with social cognition penalties and a spatial-temporal precognition module for auxiliary learning. We briefly review these components before introducing our contribution.

\subsubsection{Main Policy Architecture}

The policy network accepts depth imagery and relative goal coordinates as input at each timestep, outputting navigation actions. Visual features are extracted via a ResNet-50~\cite{resnet} encoder, while goal information passes through a linear transformation. A 2-layer LSTM~\cite{lstm} processes temporal dependencies, with outputs feeding into actor and critic heads for action generation and value estimation respectively.

The training objective combines standard PointGoal navigation rewards with social cognition penalties. The base reward at timestep $t$ follows:

\begin{equation}
R^{\hspace{1pt}t}_\text{base} = -\beta_d\Delta_d - r_{\text{slack}} + \beta_{\text{succ}} \cdot I_{\text{succ}}.
\end{equation}

where $\Delta_d$ represents geodesic distance change, $r_{\text{slack}}$ penalizes unnecessary actions, $I_{\text{succ}}$ indicates successful navigation, and $\beta_d$, $\beta_{\text{succ}}$ are weighting coefficients.

To encourage social compliance, Falcon introduces three penalty terms:

\textbf{Collision Penalty} discourages contact with obstacles and humans:
\begin{equation}
    r_{\text{coll}} = \beta_s \cdot I_{\text{s\_coll}} + \beta_h \cdot I_{\text{h\_coll}}.
\end{equation}

\textbf{Proximity Penalty} maintains safe interpersonal distances:
\begin{equation}
r_{\text{prox}} = \sum_{i=1}^{N} 
\begin{cases} 
\beta_{\text{prox}} \cdot \exp\left( -d_i^t\right) & \text{if } d_i^t < 2.0 \mathrm{~m}, \\
0 & \text{otherwise}.
\end{cases}
\end{equation}
where $d_i^t = \lVert \tau_a(t) - \tau_i(t) \rVert$ measures agent-human distance.

\textbf{Path Blocking Penalty} prevents obstruction of predicted human trajectories:
\begin{equation}
r_{\text{path}} = \sum_{k=t+1}^{t+H} \sum_{i=1}^{N} 
\begin{cases} 
\beta_{\text{path}} \cdot \left( \frac{1}{k - t + 1} \right) & \text{if } d_{\text{path}\_i}^k < 0.05 \mathrm{~m}, \\
0 & \text{otherwise.}
\end{cases}
\end{equation}

The complete reward function integrates these components:
\begin{equation}
R^{\hspace{1pt}t}_\text{total} = R^{\hspace{1pt}t}_\text{base} - (r_{\text{coll}} + r_{\text{prox}} + r_{\text{path}}).
\end{equation}

\subsubsection{Auxiliary Task Learning}

Falcon employs three auxiliary tasks to enhance spatial-temporal understanding:

\textbf{Population Estimation} predicts the number of humans present, formulated as classification over $\{0, 1, \dots, M\}$ with Cross-Entropy loss:
\begin{equation} 
\mathcal{L}_{\text{count}} = - \sum_{k=0}^{M} n_k \log(\hat{n}_k).
\end{equation}

\textbf{Position Estimation} regresses current human locations relative to the agent using MSE loss:
\begin{equation}
    \mathcal{L}_{\text{pos}} = \frac{1}{|\mathcal{M}|} \sum_{i \in \mathcal{M}} \|\hat{\mathbf{P}}_{i}^{\hspace{1pt}t} - \mathbf{P}_{i}^{\hspace{1pt}t}\|^2.
\end{equation}

\textbf{Trajectory Forecasting} predicts future positions over $H$ timesteps:
\begin{equation}
\mathcal{L}_{\text{traj}} = \frac{1}{|\mathcal{M}|} \sum_{i \in \mathcal{M}} \|\hat{\mathbf{P}}_{i}^{\hspace{1pt}t+1:t+H} - \mathbf{P}_{i}^{\hspace{1pt}t+1:t+H}\|^2,
\end{equation}
where $\mathcal{M}$ is a validity mask handling human counts.

\subsection{Proactive Risk Perception Module}

While Falcon effectively predicts human trajectories and penalizes path blocking, we identify an opportunity to enhance collision avoidance through explicit risk quantification. Our \textbf{Proactive Risk Perception Module} introduces a distance-based risk assessment mechanism that complements trajectory prediction with continuous risk scoring.

\subsubsection{Motivation}

Our risk perception module addresses two key limitations: (1) while trajectory prediction anticipates future positions, it does not explicitly quantify collision danger; (2) social cognition penalties provide sparse learning signals only when violations occur. By introducing continuous, distance-based risk assessment, we provide dense supervisory signals that guide the agent toward proactive avoidance behaviors even before entering penalty zones.
This approach enables the agent to develop more nuanced spatial reasoning, learning to maintain comfortable margins around humans rather than merely avoiding contact. The graduated risk formulation mirrors human navigation intuition, where proximity perception naturally guides movement decisions.

\subsubsection{Architecture Design}

The risk perception module consists of a lightweight neural network that processes the LSTM hidden state $\delta_R$ from the main policy network:

\begin{equation}
\phi_{\text{risk}}(\delta_R) = \sigma(W_2 \cdot \text{ReLU}(W_1 \cdot \delta_R))~,
\end{equation}
where $W_1 \in \mathbb{R}^{h \times h}$ and $W_2 \in \mathbb{R}^{M \times h}$ are learnable weight matrices, $h$ denotes hidden dimension, and $\sigma$ is the sigmoid activation producing risk scores $\hat{r}_i \in [0, 1]$ for each potential human $i \in \{1, \dots, M\}$.

\subsubsection{Risk Score Formulation}

We define ground-truth risk scores based on agent-human distances, incorporating three graduated risk zones:

\begin{equation}
r_i^{\text{true}} = 
\begin{cases}
1.0 & \text{if } d_i < d_{\text{danger}}, \\
\frac{d_{\text{safe}} - d_i}{d_{\text{safe}} - d_{\text{danger}}} & \text{if } d_{\text{danger}} \leq d_i < d_{\text{safe}}, \\
0.0 & \text{if } d_i \geq d_{\text{safe}},
\end{cases}
\end{equation}
where $d_{\text{danger}} = 2.0$ m defines the proximity threshold, $d_{\text{safe}} = 4.0$ m establishes the safe distance boundary, and $d_i = \|\tau_a(t) - \tau_i(t)\|$ measures instantaneous separation.

This formulation creates three distinct behavioral zones: a danger zone requiring immediate avoidance, a warning zone encouraging cautious navigation, and a safe zone permitting normal operation.

\subsubsection{Integration with Falcon}

The risk perception module operates in parallel with Falcon's auxiliary tasks. The total training objective becomes:

\begin{equation}
\mathcal{L}_{\text{total}} = \beta_{\text{main}} \mathcal{L}_{\text{main}} + \beta_{\text{aux}} (\mathcal{L}_{\text{count}} + \mathcal{L}_{\text{pos}} + \mathcal{L}_{\text{traj}}) + \beta_{\text{risk}} \mathcal{L}_{\text{risk}},
\end{equation}
where $\beta_{\text{risk}}$ controls the risk module's influence. In our experiments, we set $\beta_{\text{risk}} = 0.1$ to balance risk awareness with other learning objectives.

Crucially, the risk module shares the same encoded state representation $\delta_R$ as Falcon's auxiliary tasks, enabling efficient multi-task learning without additional computational overhead during inference. The module operates exclusively during training, with learned risk awareness implicitly encoded into the main policy's behavior.

%% file: sections/4_experiments.tex
\section{Experiments}
\label{sec:experiments}

In this section, we first introduce the dataset and evaluation metrics used in the IROS 2025 RoboSense Challenge Social Navigation Track. We then present our experimental setups and report the competition results, comparing our method with other participating teams' methods.

\subsection{Dataset}

We use the official data provided by the \textit{RoboSense Challenge 2025} \cite{robosense_challenge_2025} held at IROS 2025. This competition builds upon the legacy of the \textit{RoboDepth Challenge 2023} \cite{robodepth_challenge_2023,kong2023robodepth} at ICRA 2023 and the \textit{RoboDrive Challenge 2024} \cite{robodrive_challenge_2024,xie2025robobev} at ICRA 2024, continuing the collective effort to advance robust and scalable robot perception. Each track in this competition is grounded on an established benchmark designed for evaluating real-world robustness and generalization \cite{xie2025drivebench,gong2025cognition,li2024place3d,chu2024geotext-1652,liang2025pi3det}. 

Specifically, this task is built upon the \textbf{Falcon} benchmark \cite{gong2025cognition} in \textbf{Track 2}, which assesses socially compliant navigation policies in dynamic human environments using realistic RGB-D simulations and human-robot interaction metrics. This task employs the Social-HM3D benchmark~\cite{gong2025cognition}, a large-scale photo-realistic dataset specifically designed for evaluating socially-aware navigation in dynamic indoor environments. This benchmark addresses critical limitations in existing datasets by providing diverse scene types with balanced human density and naturalistic agent behaviors.

\subsubsection{Dataset Characteristics}
Social-HM3D is constructed from 3D-reconstructed real-world indoor scenes sourced from the HM3D dataset~\cite{ramakrishnan2021habitat}, featuring 844 unique environments spanning residential, commercial, and public spaces. The dataset incorporates collision-aware human agents with goal-directed behaviors, enabling realistic evaluation of human-robot interaction dynamics. Key design principles include:

\textit{Goal-Oriented Trajectories:} Human agents navigate between designated waypoints within each scene, creating natural movement patterns that vary in complexity based on environmental layout. This design contrasts with prior datasets that employ random walk behaviors~\cite{igibsonchallenge2021,vuong2023habicrowd}, providing more authentic interaction scenarios.

\textit{Naturalistic Motion Dynamics:} Walking speeds are randomized within 0.8-1.2 times the robot's velocity to reflect human variability. Agents employ mutual collision avoidance using the ORCA algorithm~\cite{van2011reciprocal}, alternating between locomotion and stationary periods based on task completion. Realistic skeletal animations enhance visual fidelity and temporal coherence.

\begin{figure}[t]
    \centering
    \includegraphics[width=1.0\linewidth]{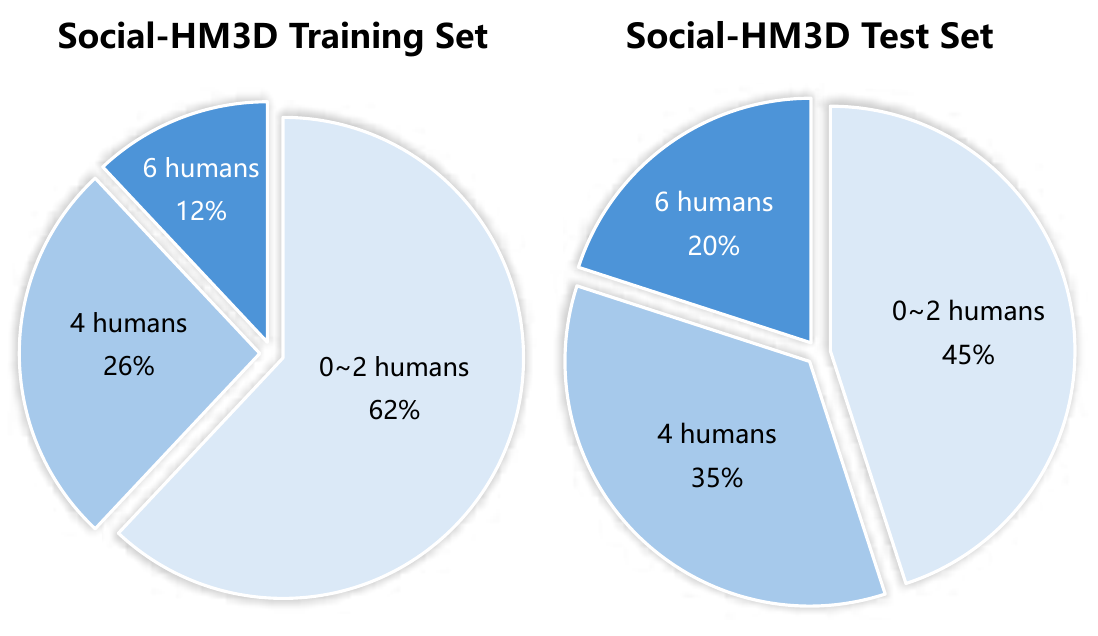}
    \caption{Human Distribution by Scene Area in Social-HM3D (Train/Test Set): The benchmark calibrates human density based on diverse scene areas. Scenes are categorized: small spaces (0-40 m$^2$) with 0-2 humans, medium spaces (40-80 m$^2$) with 4 humans, and large spaces ($> 80$ m$^2$) with 6 humans. This area-proportional scaling ensures realistic social interaction density while preventing overcrowding in human-shared environments.}
\label{fig:Social-HM3D_density}
\end{figure}

\textit{Calibrated Population Density:} Human counts are scaled proportionally to scene area, preventing both sparse environments with insufficient interaction opportunities and overcrowded spaces that constrain movement. As illustrated in Fig.~\ref{fig:Social-HM3D_density}, the dataset employs diverse density categories. This calibration ensures consistent challenge difficulty across diverse spatial configurations.

\subsubsection{Challenge Requirements \& Restrictions}
Participants train navigation policies on the \texttt{social-hm3d/train} split, which provides access to scene geometry, human trajectories, and RGBD observations. During development, teams can leverage the \texttt{social-hm3d/val} split approximately 1000 episodes for local validation, enabling iterative algorithm refinement without accessing the held-out test data. The final evaluation is conducted on a private test set comprising approximately 500 episodes from previously unseen scenes in \texttt{social-hm3d/test}. This held-out evaluation protocol rigorously assesses generalization capabilities without scene-specific overfitting.

Agents operate under egocentric constraints, receiving only depth imagery, odometry data, and relative goal coordinates. Global maps, human position oracles, and privileged environmental information are explicitly prohibited, ensuring realistic deployment conditions. The benchmark emphasizes socially compliant behaviors including personal space maintenance, collision avoidance, and efficient goal-reaching under dynamic human presence.

\begin{table*}[ht!]
\centering
\resizebox{\textwidth}{!}{
\begin{tabular}{c|l|ccccc}
\toprule
\textbf{Rank} & \textbf{Participant Team} & \textbf{SR ($\uparrow$)} & \textbf{SPL ($\uparrow$)} & \textbf{PSC ($\uparrow$)} & \textbf{H-Coll ($\downarrow$)} & \textbf{Total ($\uparrow$)} \\
\midrule
1 & Are Ivan & 0.6600 &0.5977 & 0.8629 & 0.3240 & 0.7022 \\
\rowcolor{yellow!30}
\textbf{2} & \textbf{Xiaomi EV-AD VLA (Ours)} & \textbf{0.6560} & \textbf{0.5958} & \textbf{0.8608} & \textbf{0.3300} & \textbf{0.6994} \\
3 & Auto Robot & 0.6480 & 0.6010 & 0.8607 & 0.3420 & 0.6977 \\
4 & DUO & 0.6520 & 0.5855 & 0.8611 & 0.3260 & 0.6948 \\
5 & CityU-ASL & 0.6440 & 0.5951 & 0.8582 & 0.3280 & 0.6936 \\
6 & Wrong Team & 0.6480 & 0.5865 & 0.8591 & 0.3300 & 0.6929 \\
7 & diting & 0.6240 & 0.5787 & 0.8626 & 0.3320 & 0.6820 \\
8 & zhengxinhan & 0.6260 & 0.5752 & 0.8626 & 0.3540 & 0.6817 \\
9 & Social Dog & 0.6320 & 0.5660 & 0.8613 & 0.3420 & 0.6810 \\
10 & MW & 0.6220 & 0.5520 & 0.8691 & 0.3080 & 0.6751 \\
11 & Dog of social & 0.6080 & 0.5478 & 0.8637 & 0.3700 & 0.6666 \\
12 & Path Seekers & 0.5880 & 0.5507 & 0.8662 & 0.3520 & 0.6603 \\
13 & GRAM & 0.5680 & 0.5057 & 0.8663 & 0.3800 & 0.6388 \\
14 & CORE Lab & 0.5400 & 0.4997 & 0.8630 & 0.3920 & 0.6248 \\
15 & 1111111212312312 & 0.5400 & 0.4997 & 0.8630 & 0.3920 & 0.6248 \\
16 & Falcon (baseline) & 0.5400 & 0.4997 & 0.8630 & 0.3920 & 0.6248 \\
\bottomrule
\end{tabular}
}
\caption{\textbf{Competition Results on IROS 2025 RoboSense Challenge Social Navigation Track.} The Falcon serves as the official baseline. Our team (Xiaomi EV-AD VLA) achieved 2nd place among 16 participating teams with our results \colorbox{yellow!30}{highlighted}.}
\label{tab:competition_results}
\vspace{-10pt}
\end{table*}

\subsection{Evaluation Metrics}
The benchmark assesses performance across two principal dimensions: task completion efficiency and social compliance. The specific metrics are defined as follows:

\begin{itemize}
    \item \textbf{Success Rate (SR):} Percentage of episodes where the agent reaches within 1.0 meter of the goal position.
    
    \item \textbf{Success weighted by Path Length (SPL):} Evaluates path efficiency relative to the optimal trajectory:
    \begin{equation}
    \text{SPL} = \frac{1}{N} \sum_{i=1}^{N} S_i \cdot \frac{l_i}{\max(p_i, l_i)}.
    \end{equation}
    where $l_i$ denotes the shortest path length, $p_i$ represents the actual path taken, and $S_i$ indicates success ($1$ if successful, $0$ otherwise).
    
    \item \textbf{Personal Space Compliance (PSC):} Computed as the percentage of timesteps maintaining at least 0.5 meters separation from all humans, reflecting socially appropriate spacing behavior.
    
    \item \textbf{Human Collision Rate (H-Coll):} Percentage of episodes involving human collisions. Collisions result in episode failure and negatively impact both SR and PSC.
    
    \item \textbf{Total Score:} A weighted aggregate metric that determines final rankings, computed as:
    \begin{equation}
    \text{Total} = 0.4 \times \text{SR} + 0.3 \times \text{SPL} + 0.3 \times \text{PSC}.
    \end{equation}
    This formulation emphasizes task success while balancing path efficiency and social compliance. Human collisions are implicitly penalized through their negative impact on SR and PSC.
\end{itemize}

\subsection{Experiment Setups}

Our implementation builds upon the Falcon framework with the integrated Proactive Risk Perception Module. The visual encoder employs ResNet-50~\cite{resnet} architecture for depth processing, while a 2-layer LSTM handles temporal feature extraction. The risk perception module adds minimal computational overhead with two fully-connected layers operating on the shared hidden state representation.

We adopt the DD-PPO algorithm~\cite{wijmans2019dd} for policy optimization. The model is initialized from pretrained PointNav weights provided by the Habitat platform~\cite{ramakrishnan2021habitat} and fine-tuned on the social navigation task. Training is conducted on 4 NVIDIA A40 GPUs with 8 parallel environments for approximately 75 million steps.

\subsection{Competition Results}

Table~\ref{tab:competition_results} presents the final leaderboard rankings on the private test set. Our submission (Team Xiaomi EV-AD VLA) achieved second place among 16 participating teams with a total score of 0.6994, demonstrating strong performance across all evaluation metrics. Our method achieves 0.656 SR and 0.5958 SPL, indicating effective goal-reaching capability with reasonably efficient paths. The 0.8608 PSC demonstrates strong adherence to social distancing norms, while the 0.33 H-Coll reflects the challenge's difficulty in crowded scenarios with dynamic human agents.

Compared to the top-performing team (0.7022 total score), our method shows competitive performance with only a 0.0028 difference. This narrow margin suggests our approach successfully enhances navigation safety while maintaining task efficiency. The results validate our hypothesis that explicit risk quantification complements trajectory prediction for improved collision avoidance.

\begin{figure}[ht!]
    \centering
    \includegraphics[width=1.0\linewidth]{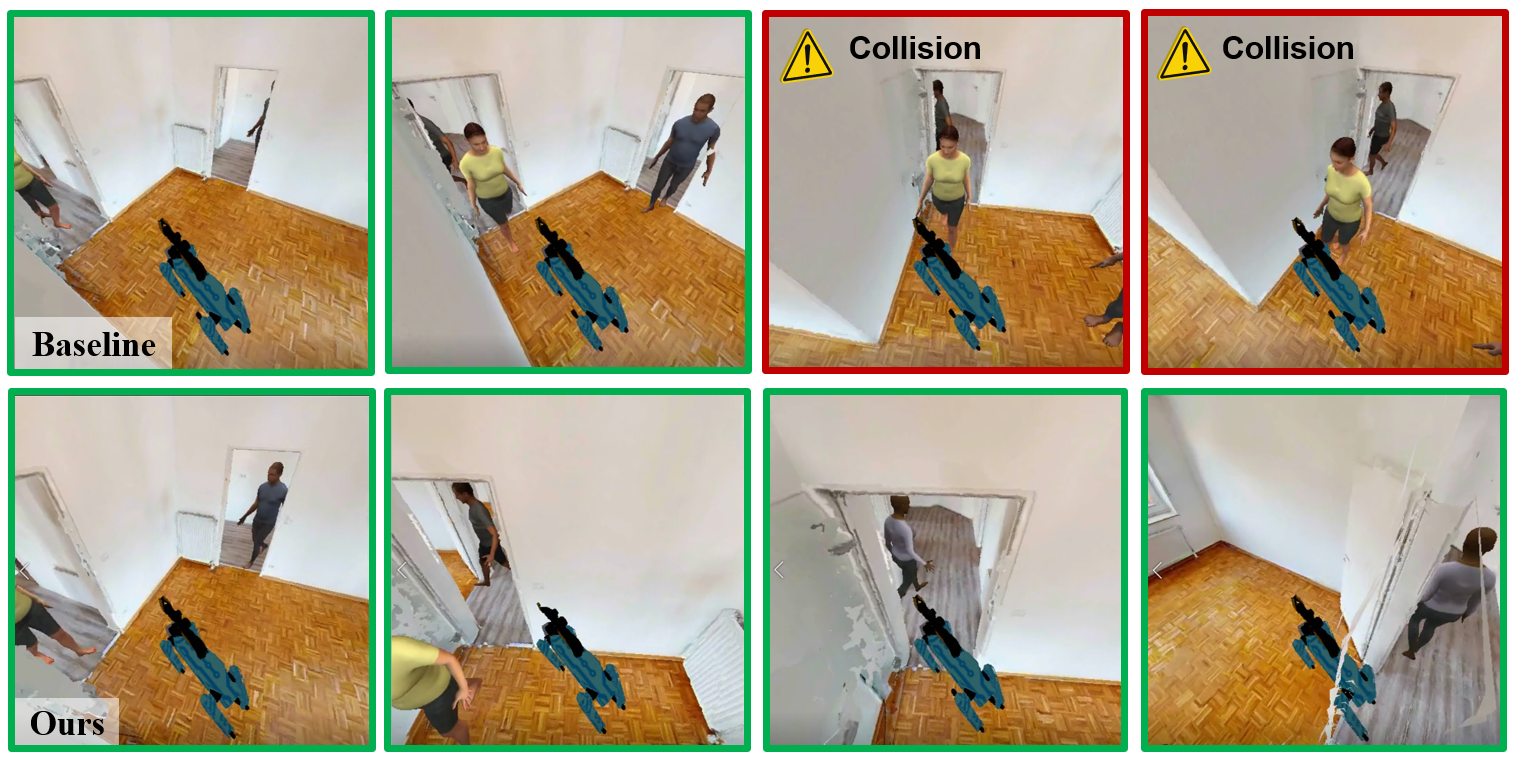}
    \caption{\textbf{Episode demonstration.} Compared to baseline, our method successfully predicts human trajectories, proactively moves to a non-obstructive position, and then avoids collisions. Green indicates safe behaviors, and red indicates collisions with humans.}
    \label{fig:example}
    \vspace{-15pt}
\end{figure}

%% file: sections/5_conclusion.tex
\section{Conclusion}
\label{sec:conclusion}



In this report, we present our technical submission to the IROS 2025 RoboSense Challenge Social Navigation Track. Building upon the Falcon framework~\cite{gong2025cognition}, we introduce a Proactive Risk Perception Module that enhances collision avoidance capabilities. Our comprehensive evaluation on the Social-HM3D benchmark demonstrates the effectiveness of risk-aware auxiliary learning for social navigation. The competition results validate our approach, achieving a competitive 2nd place among 16 participating teams.